\documentclass[sigconf,nonacm,natbib=false]{acmart}

\usepackage{algorithmic}
\usepackage{algorithm}
\usepackage{amsfonts}
\usepackage{amsmath}
\usepackage{array}
\usepackage{bm}
\usepackage{booktabs}
\usepackage[outline]{contour}
\usepackage[inline]{enumitem}
\usepackage{fontawesome5}
\usepackage{float}
\usepackage{graphicx}
\usepackage{hyperref}
\usepackage{makecell}
\usepackage{listings}
\usepackage{placeins}
\usepackage[eulergreek]{sansmath}
\usepackage{siunitx}
\usepackage{subcaption}
\usepackage{stfloats}
\usepackage{textcomp}
\usepackage{tikz}
\usepackage{url}
\usepackage{verbatim}
\usepackage{xcolor}
\usepackage{cleveref}

\usepackage{hyperref}
\usepackage{doi}

\usetikzlibrary{decorations.pathreplacing, arrows}
\usetikzlibrary{positioning, arrows.meta, calc,fit}
\definecolor{color0}{RGB}{70,100,170} 
\definecolor{color1}{RGB}{252,229,0} 
\definecolor{color2}{RGB}{35,161,224} 
\definecolor{color3}{RGB}{162,34,35} 
\definecolor{color4}{RGB}{0,150,130} 
\definecolor{color5}{RGB}{140,182,60} 
\definecolor{color6}{RGB}{163,16,124} 
\definecolor{color7}{RGB}{167,130,46} 
\setcopyright{acmlicensed}
\copyrightyear{2018}
\acmYear{2018}
\acmDOI{XXXXXXX.XXXXXXX}
\acmConference[SC25]{Sustainable Supercomputing Workshop}{November 16--21 2025}{St. Louis, MO}

\acmISBN{978-1-4503-XXXX-X/2018/06}

\begin{document}

\title{Energy Consumption in Parallel Neural Network Training}


\author{Philipp Huber}
\affiliation{
  \institution{Karlsruhe Institute of Technology}
  \city{Karlsruhe}
  \country{Germany}}
\email{philipp.huber@kit.edu}
\orcid{0000-0001-6862-8177}

\author{David Li}
\affiliation{
  \institution{Karlsruhe Institute of Technology}
  \city{Karlsruhe}
  \country{Germany}}
\email{david.li@kit.edu}
\orcid{0009-0005-5057-5181}

\author{Juan Pedro Gutiérrez Hermosillo Muriedas}
\affiliation{
  \institution{Karlsruhe Institute of Technology}
  \city{Karlsruhe}
  \country{Germany}}
\email{juan.muriedas@kit.edu}
\orcid{0000-0001-8439-7145}

\author{Deifilia Kieckhefen}
\affiliation{
  \institution{Karlsruhe Institute of Technology}
  \city{Karlsruhe}
  \country{Germany}}
\email{deifilia.to@kit.edu}
\orcid{0009-0000-1197-5493}

\author{Markus Götz}
\affiliation{
  \institution{Karlsruhe Institute of Technology}
  \city{Karlsruhe}
  \country{Germany}}
\email{markus.goetz@kit.edu}
\orcid{0000-0002-2233-1041}

\author{Achim Streit}
\affiliation{
  \institution{Karlsruhe Institute of Technology}
  \city{Karlsruhe}
  \country{Germany}}
\email{achim.streit@kit.edu}
\orcid{0000-0002-5065-469X}

\author{Charlotte Debus}
\affiliation{
  \institution{Karlsruhe Institute of Technology}
  \city{Karlsruhe}
  \country{Germany}}
\email{charlotte.debus@kit.edu}
\orcid{0000-0002-7156-2022}

\renewcommand{\shortauthors}{Huber et al.}

\begin{abstract}
The increasing demand for computational resources of training neural networks leads to a concerning growth in energy consumption.
While parallelization has enabled upscaling model and dataset sizes and accelerated training, its impact on energy consumption is often overlooked.
To close this research gap, we conducted scaling experiments for data-parallel training of two models, ResNet50 and FourCastNet, and evaluated the impact of parallelization parameters, i.e., GPU count, global batch size, and local batch size, on predictive performance, training time, and energy consumption.
We show that energy consumption scales approximately linearly with the consumed resources, i.e., GPU hours; however, the respective scaling factor differs substantially between distinct model trainings and hardware, and is systematically influenced by the number of samples and gradient updates per GPU hour.
Our results shed light on the complex interplay of scaling up neural network training and can inform future developments towards more sustainable AI research.

\end{abstract}

\ccsdesc[500]{Computing methodologies~Neural networks}
\ccsdesc[500]{Hardware~Power and energy}
\ccsdesc[500]{Computing methodologies~Parallel computing methodologies}

\keywords{GreenAI, Energy Efficiency, HPC in AI, Distributed Neural Networks, Data Parallel, deep learning, Scaling}


\maketitle

\section{Introduction}
Modern deep learning (DL) is fundamentally coupled to high-per\-formance computing (HPC).
In order to achieve higher prediction accuracy, both dataset and model sizes are growing rapidly.
The resulting demand for ever more computational resources necessitates the use of parallel training approaches~\cite{ben2019demystifying}.
Among the different varieties of parallel DL, \emph{data parallelism} (DP) is by far the most common one~\cite{keuper2016distributed}.
It reduces training time effectively by distributing subsets of the dataset across multiple GPUs.
In each training step, each GPU holds the same copy of the model and performs the forward--backward pass on the local dataset.
Communication is only required at the end of each forward--backward pass to synchronize gradients and can be efficiently overlaid with computation~\cite{li2020pytorchdistributed}. 
However, scaling to dozens or hundreds of GPUs introduces some challenges:
When keeping the per-GPU number of samples in a gradient update (local batch size) fixed, adding more and more GPUs diminishes prediction accuracy at some point. 
This effect is known as \emph{large batch effects}~\cite{goyal2017accurate} and ultimately limits the scalability of data-parallel training.
Though means and methods to alleviate large batch effects to some extent have been proposed~\cite{you2017large,yamazaki2019yet,malladi2022sdes}, they remain a challenge. 
Conversely, keeping the global number of samples within a gradient update (global batch size) constant reduces the per-GPU workload at larger numbers of GPUs, eventually leading to inefficiencies and communication barriers.

One aspect that is typically overlooked in scaling data-parallel DL to large GPU counts is the associated energy consumption. 
Recent efforts have raised awareness on the topic of energy efficiency and Green AI\cite{debus2023reporting}, in particular for large language models (LLMs)\cite{dauner2025energy}. 
Yet, it is unclear which role parallelism and scalability play in the energy consumption of neural network training.
On the one hand, parallelism is required to tackle the increasingly large datasets and to keep training time feasible.
On the other hand, scaling can result in disproportional energy consumption and decreased model performance.
Efficient scaling of model training to multi-GPU systems requires balancing this delicate trade-off between accuracy, training time, and energy efficiency.
However, the exact interplay between these three axes in DL parallelism and scalability has scarcely been studied and is consequently far from understood.

Our study aims to shed light on the energy and performance cost of scaling up DL model training for the purpose of reducing training time.
In a set of scaling experiments for typical training workloads, we vary parallelization setups and evaluate the differences in energy consumption, overall training time, and final model prediction accuracy. 
Our contributions include:
\begin{itemize}
    \item In-depth scaling experiments of training ResNet50~\cite{he2016deep} on the ImageNet-2012 dataset~\cite{deng2009imagenet}, using both fixed and varying dataset sizes, evaluating each with both constant global and local batch.
    \item In-depth scaling experiments of training FourCastNet~\cite{pathak2022fourcastnet, fcn_code} on the ERA5 dataset~\cite{hersbach2020era5}, using a fixed dataset size evaluating both constant global and local batch sizes.
    \item Detailed analysis of the interplay between energy consumption, training time, and model accuracy, including GPU power profiles and different hardware setups.
\end{itemize}

\section{Related Work}

The immense energy consumption and associated carbon footprint of AI workloads, and in particular neural network (NN) training, have recently drawn an increasing amount of research interest~\cite{schwartz2020green,luccioni2019quantifying, debus2023reporting}.
Several studies have benchmarked different aspects of energy consumption and efficiency of training deep NNs~\cite{mehlin2023towards}.
For example, the trade-off between accuracy and energy efficiency of training convolutional NNs for image classification~\cite{xu2023energy} and the relationship between dataset size, network structure, and energy use for fully connected NNs~\cite{tripp2024measuring} have been studied.
Furthermore, the trade-off between energy consumption and hardware performance for recurrent NNs~\cite{you2023zeus} and the energy consumption of different NN-workloads on different hardware configurations~\cite{caspart2022precise} were investigated.

Next to architectural and hardware choices, the impact of hyperparameter optimization on energy consumption has been studied in-depth~\cite {yarally2023uncovering, geissler2024spend}.
In particular, Gei{\ss}ler et al.~\cite{geissler2024power} studied the impact of hyperparameters, such as learning rate, batch size, and knowledge transfer techniques, on energy consumption across different hardware systems. 
However, the above-mentioned studies focused on small datasets and models that can be trained on single-GPU setups.
Frey et al.~\cite{frey2022benchmarking} conducted large-scale experiments on neural networks for processing, computer vision, and chemistry using up to 424 GPUs, analyzing power utilization, GPU clock rate limits, and training time scaling under compute and energy constraints.
Kozczal et al.~\cite{koszczal2023performance} exploited power capping strategies for balancing scalability and energy consumption in multi-GPU systems up to eight GPUs.

With growing sizes of datasets, data-parallel training is by now a fundamental and essential part of NN training~\cite{liang2024communication}.
Scalability of DP has been intensely studied, primarily aiming to reduce training time by circumventing communication bottlenecks.
A central objective is to improve predictive performance without increasing computational cost~\cite{shen2024efficient}.
Several techniques have been proposed to mitigate the communication bottleneck, including gradient accumulation, topology-aware communication patterns, asynchronous methods, and gradient compression~\cite{vogels_powersgd_2019,abrahamyan2022compression}. 
Approaches using a parameter server for aggregating gradients often rely on asynchronous communication, though they suffer from stale gradients~\cite{niu_hogwild_2011}.
Methods such as hierarchical local SGD~\cite{lin_dont_2020} and DASO~\cite{coquelin2022accelerating} reduce communication overhead by localizing synchronization within smaller subgroups between global updates, leveraging network topology to accelerate training.
Ahn et al.~\cite{ahn2024efficient} proposed a method to improve the efficiency of DP by tackling heterogeneous hardware architectures. 
Though data-parallel training has become the de facto standard setup for training NNs on large dataset, its scalability is limited not only by network communication, but by model accuracy itself. 
It is known to introduce large batch effects, i.e., diminishing prediction accuracy beyond a certain global batch size as observed for ResNet on ImageNet~\cite{goyal2017accurate} or for Vision Transformers~\cite{chen2021empirical}.

\section{Energy Costs of Data-Parallel Neural Networks}
We conducted scaling experiments with varying parallelization parameters, i.e., local batch size (LBS), global batch size (GBS), and number of GPUs, for training two models to investigate the balance between prediction accuracy, training time, and energy consumption.
As a first model, we studied ResNet~\cite{he2016deep} on the ImageNet-2012 dataset~\cite{deng2009imagenet} for image classification, which is a paradigm for accelerating NN training through data parallelism.
As the second model, we examined FourCastNet~\cite{pathak2022fourcastnet}, a weather forecasting model based on a Vision Transformer using adaptive Fourier neural operators as a backbone.
FourCastNet is trained on 20 atmospheric variables at 0.25$^\circ$ resolution of the ERA5 global reanalysis dataset~\cite{hersbach2020era5}. 
Compared to ResNet50 trained on ImageNet, it provides a substantially more compute-intensive workload regarding both model and dataset size.
While for both models, training time and accuracy scaling have been studied~\cite{kurth2023fourcastnet, goyal2017accurate}, energy consumption under varying parallelization parameters remains unexplored.

\begin{figure*}
    \centering
    \input{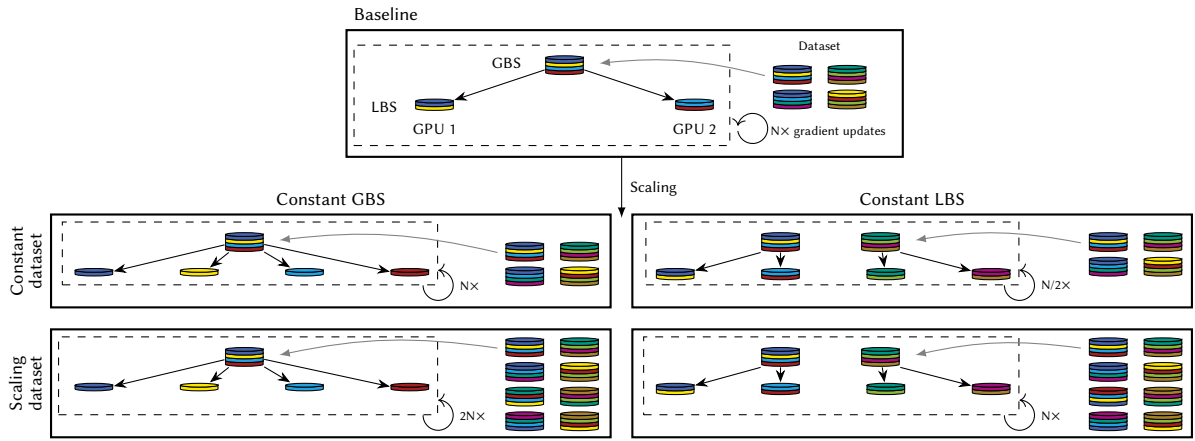}
    \caption{Graphical overview of scaling experiments.
    Starting from a baseline configuration of parallelization parameters, four different scaling procedures are studied by increasing the GPU count.
    The global batch size (GBS), local batch size (LBS), or dataset size are scaled or kept constant as indicated.
    Depending on the distinct scaling, the number of gradient updates $N$ changes.}
    \Description{}
    \label{fig:experimental_design}
\end{figure*}

Our scaling experiments use either a fixed dataset or scale the samples proportionally with the GPU count.
Both of these setups can be run with either a fixed per gradient-update workload per GPU (constant LBS) or a decreasing per gradient-update workload per GPU (constant GBS).  
Thus, a total \emph{workload} can be defined as $W=N_\mathrm{gradient\ updates}\cdot n_\mathrm{GPUs}\cdot\mathrm{LBS}$.
In \Cref{fig:experimental_design}, we provide an explanatory overview of these variants.

\subsection{Hardware Specifications and Software Stack}
\label{sec:hardware}
All experiments were executed on the HoreKa supercomputing system, which is equipped
with NVIDIA A100-40GB and NVIDIA H100-94GB nodes, connected via 4X HDR 200 GBit/s InfiniBand interconnect.
Each node contains four GPUs with two CPU sockets.
For the A100 nodes, each CPU socket consists of 76 Intel Xeon Platinum 8368 cores. 
For the H100 nodes, each CPU socket consists of 64 AMD EPYC 9354 cores.
We used Python 3.11, OpenMPI 4.1, and CUDA 12.2 (ResNet) or CUDA 12.6 (FourCastNet), with \texttt{torch 2.7.0}, \texttt{torchvision} 0.22.0, and \texttt{mpi4py} 4.0.3.
Both models are trained using \texttt{DistributedDataParallel} from \texttt{torch}.
Power draw and energy consumption were measured using the Python package \texttt{perun}~\cite{gutierrez2023perun} with a sampling rate of \SI{1}{\second}, which supports multi-node processing using the Message Passing Interface (MPI).
We define the total energy as the sum of GPU, CPU, and RAM energy. 
The share of the RAM energy to the total energy is low ($ \leq \mathrm{6} \%$) throughout all experiments.

\subsection{ResNet}
\label{sec:resnet}

\subsubsection{Setup}
We used the \texttt{torchvision} implementation of ResNet50, consisting of 50 residual blocks with three layers each, together, and trained it on ImageNet-2012, \num{1281167} training samples and \num{50000} validation samples across \num{1000} classes\footnote{Code will be made available upon publication.}.
A standard image preprocessing and a warm-up for the learning rate of five epochs as proposed by Goyal et al.\cite{goyal2017accurate} was applied.
After a warm-up, a \texttt{ReduceOnPlateau} scheduler was used with a factor of \num{0.5} and a patience of five.
Similar to He et al.~\cite{he2016deep}, we used a weight decay of \num{0.0001} and a momentum of \num{0.9} for the stochastic gradient descent optimizer.
We used four workers in the dataloader and applied \texttt{CrossEntropy} for the loss function.
The model was trained for 100 epochs. 
The goal is not to achieve optimal prediction accuracy, but instead to highlight the differences arising from distributing the batch across multiple GPUs. 
Prediction accuracy in terms of top-1 error, i.e., the accuracy with which the model predicts the image labels correctly with a single attempt, was computed on the validation set.
Each experiment was conducted five times with the same random seed to account for fluctuations in hardware performance, which leads to deviations in training time and energy consumption, and the root mean square error (RMSE) was computed.
In experiments utilizing fewer than four GPUs per node, the full capacity of both CPU sockets was utilized.

\subsubsection{Scaling with Constant Dataset Size}
\label{sec:resnet_cDS}

\begin{figure*}
  \centering
  \includegraphics[width=\linewidth]{ 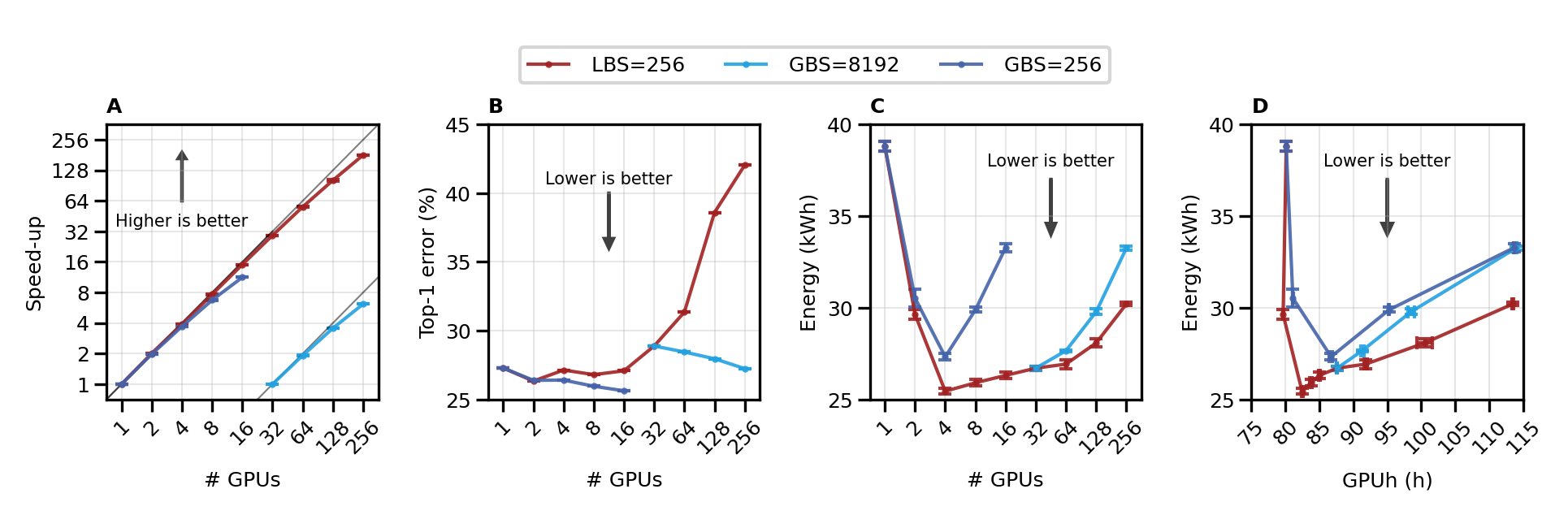}
  \caption{Scaling experiments for ResNet50 trained on the complete ImageNet-2012 dataset.
  Speed-up (A), top-1 error (B), and energy consumption (C) depending on the number of GPUs are shown.
  Additionally, the energy consumption versus GPU hours (D) is depicted.
  The number of GPUs is increased by keeping a constant local batch size (LBS) or a constant global batch size (GBS), leading to an increasing GBS or a decreasing LBS, respectively.
  The depicted error bars are determined as standard deviation over five separate runs.}
  \Description{}
  \label{fig:resnet_energy_scaling_cds}
\end{figure*}

We conducted two types of scaling experiments on a dataset of constant size, keeping either the local batch size (LBS) or the global batch size (GBS) fixed.
Results from both experiments are shown in \Cref{fig:resnet_energy_scaling_cds}.
The most common setup in data-parallel training is using a constant LBS to fully leverage the memory of each available GPU, i.e., using the largest LBS possible, and adding more GPUs, i.e, increasing the GBS.
This reduces the number of gradient updates per epoch and, consequently, accelerates training. 
However, increasing GBSs might lead to large batch effects at some point. 
We investigate this scenario by keeping the LBS=256, which is the maximum memory capacity of an A100-40GB NVIDIA GPU, and scaling the number of GPUs from $n=(\mathrm{1})$ to $n=(\mathrm{256})$.
\newline
\Cref{fig:resnet_energy_scaling_cds} shows results on the measured total training time and energy consumption of the training as well as classification quality of the trained model, i.e., top-1 accuracy. 
As expected, plain data-parallelism implemented in PyTorch's \texttt{distributed.dataparallel} provides near-optimum scaling behavior, though we observe slightly below-optimal speedup beyond 32 GPUs (cf. \Cref{fig:resnet_energy_scaling_cds}A). 
Energy consumption is, in general, continuously increasing ($n > \mathrm{4}$ GPUs) with the number of GPUs \Cref{fig:resnet_energy_scaling_cds}C).
Beyond a single node, energy consumption initially increases slowly, then rapidly beyond 32 GPUs, due to the non-linear scaling of the training time. 
For optimal speedup, the compute resources (GPU hours) remain constant, resulting in constant energy consumption.
Since the non-optimum speedup results in more GPUh, the consumed energy increases as well, following an approximately linear trend (cf. \Cref{fig:resnet_energy_scaling_cds}D).
To investigate different scaling of GPUs and CPUs, full CPU sockets were used for runs using $n \leq \mathrm{4}$ GPUs.
While this has a negligible impact on training time, it leads to a local minimum in energy consumption:
CPU energy consumption decreases from \SI{18.9}{\kilo\watt\hour} to \SI{6.7}{\kilo\watt\hour} for $n = \mathrm{1}$ to $n = \mathrm{4}$, since DP leads to shorter run times. 
However, the GPU energy increases  from \SI{17.6}{\kilo\watt\hour} to \SI{20.3}{\kilo\watt\hour} for $n = \mathrm{1}$ to $n = \mathrm{256}$ GPUs, due to inefficiencies introduced by DP.
From $n = \mathrm{1}$ to $n = \mathrm{4}$ GPUs, the share of the CPU energy to the total energy decreases from \SI{49}{\percent} to \SI{27}{\percent} , where it stays approximately constant.
Therefore, while using fewer GPUs is generally more energy efficient, decreasing the number of GPUs without decreasing the number of CPUs significantly worsens the energy efficiency.

We further observe large batch effects (cf. \Cref{fig:resnet_energy_scaling_cds}B), i.e., the onset of decreasing accuracy beyond a certain GBS, consistent with prior findings\cite{goyal2017accurate}.
Up to 32 GPUs (GBS=8\,192), the top-1 prediction accuracy remains approximately constant at about \SI{28}{\percent}, before rapidly increasing at larger batch sizes, reaching \SI{42}{\percent} for GBS=65\,536.
To circumvent these large batch effects, the GBS needs to stay below a certain threshold.
We investigate the scaling behavior of this setup by keeping the GBS fixed and decreasing the LBS at larger GPU counts.
Given that a single GPU's memory capacity limits the maximum LBS to 256, and large batch effects emerge at about GBS=8\,192, we study two regimes: a \emph{small-scale regime} of one up to 16 GPUs at GBS=256, corresponding to a lower bound of LBS=16, and a \emph{large-scale regime} of 32 to 256 GPUs at GBS=8\,192, corresponding to a lower bound of LBS=32. 
Scaling these two regimes further, i.e., to higher GPU counts, would result in very small, and thus meaningless per-GPU workloads.
We observe that keeping the GBS constant indeed circumvents the degradation in predictive performance, with a slightly worse top-1 accuracy for the large-scale regime (cf. \Cref{fig:resnet_energy_scaling_cds}B).
Training time scaling is near-optimal in both regimes (cf. \Cref{fig:resnet_energy_scaling_cds}A), with GPU hours increasing only gradually as the number of GPUs grows.
As expected, the large-scale regime has shorter training times compared to the small-scale regime.
Again, we find an energy minimum in the small-scale regime at four GPUs, due to not scaling the number of CPUs for single-node experiments.
Ignoring these values, we find that analogously to the previous experiment, both regimes exhibit a proportional correspondence between GPUh and energy consumption.
Differences in the relation of the GPUh and the energy consumption between experiments arise from the differences in workloads (see~\Cref{sec:power_scaling} for an in-depth discussion).
Given the non-ideal speed-up, energy consumption increases generally for higher numbers of GPUs.
Hence, even though utilizing higher numbers of GPUs decreases the training time, lower numbers of GPUs provide better energy efficiency.

\subsubsection{Scaling with Increasing Dataset Size}
Increasing the number of GPUs for the same dataset size, as presented before, provides a widely adopted approach in practice.
These scaling experiments ultimately encounter scalability limits, as workloads per process become too small to maintain efficiency.
To isolate scaling behavior, we conducted experiments by scaling the dataset size (both training and validation) proportionally to the number of added GPUs, keeping the overall workload per GPU constant.
Again, we study two cases, with constant LBS and with constant GBS.

\begin{figure*}
  \centering
  \includegraphics[width=\linewidth]{ 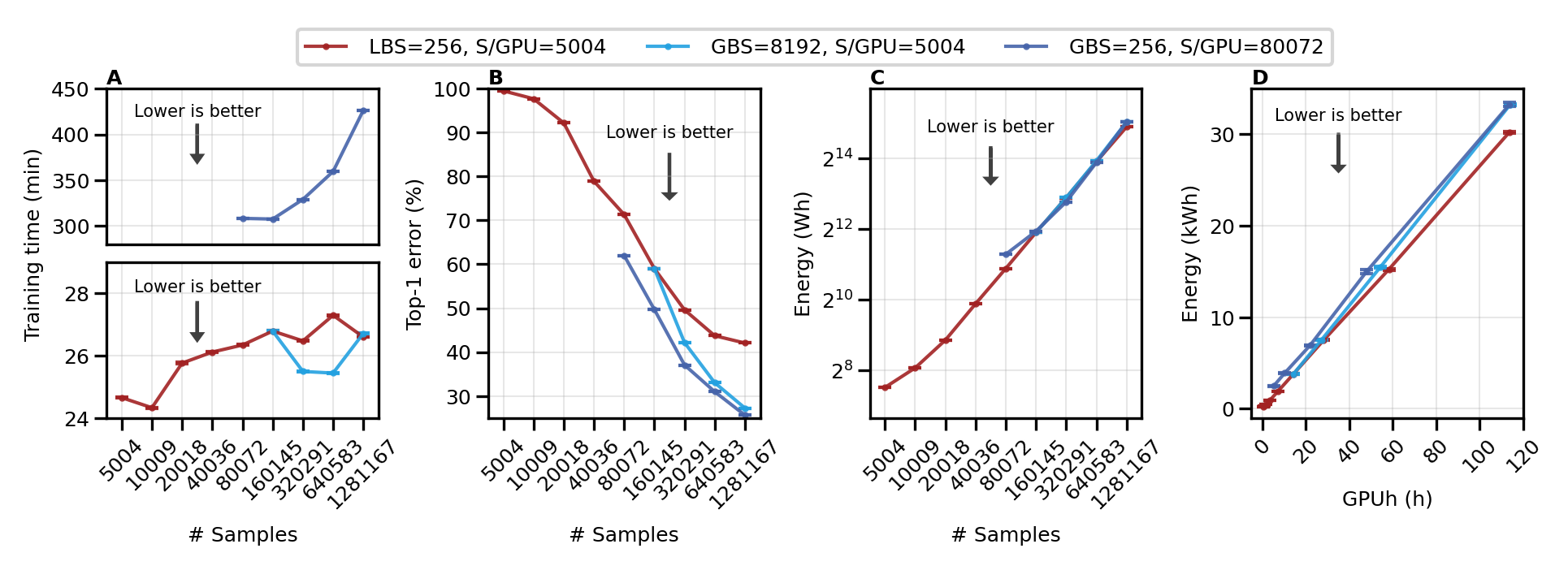}
  \caption{Scaling experiments for ResNet50 trained on the scaled ImageNet-2012 dataset.
  The overall number of samples is increased by increasing the number of GPUs, resulting in a constant number of samples per GPU (S/GPU).
  Training time (A), top-1 error (B), and energy consumption (C) depending on the number of GPUs are shown.
  Additionally, the energy consumption versus GPU hours (D) are depicted.
  The local batch size (LBS) or global batch size (GBS) are kept constant, leading to increasing GBS or decreasing LBS by increasing the number of GPUs.
  The depicted error bars are determined as the standard deviation over five separate runs.}
  \Description{}
  \label{fig:resnet_energy_scaling_vds}
\end{figure*}

Using a constant LBS, the number of gradient updates per GPU remains constant while the overall number of samples per gradient update, i.e., the GBS, increases for increasing GPU count.
This leads to counteracting effects: while training on more data generally improves prediction accuracy, larger GBS can induce large batch effects that degrade it.
We study scaling from one to 256 GPUs using a constant LBS=256 and a constant number of S/GPU=5004 samples per GPU.
As expected, the training time and, therefore, also training time efficiency remain constant (cf. \Cref{fig:resnet_energy_scaling_vds}A). 
The improvement on the top-1 error begins to stagnate at large sample counts, where large batch effects emerge (cf. \Cref{fig:resnet_energy_scaling_vds}B).
Due to constant training time, but increasing numbers of GPUs, the GPU hours are increasing, which results in increasing energy consumption (cf. \Cref{fig:resnet_energy_scaling_vds}C).
This increase is proportional to the number of samples with about  \SI{0.0235}{\watt\hour} per sample (neglecting $\mathrm{n < 4}$ GPUs) and proportional to the GPU hours (cf. \Cref{fig:resnet_energy_scaling_vds}D).

Like before, we aim to isolate the effects of large global batches. 
Hence, we conducted experiments where the GBS, i.e., the overall workload per gradient update across all GPUs, remains constant. 
Increasing the dataset size with more GPUs thus decreases the LBS, while the number of gradient updates per epoch grows.
This circumvents large batch effects, such that adding more data continually improves model accuracy.
We again study a \emph{small-scale regime} (GBS=256) with $n =\mathrm{1}$ to $n = \mathrm{16}$ GPUs and a \emph{large-scale regime} (GBS=8\,192) with $n = \mathrm{32}$ to $n = \mathrm{256}$ GPUs.
Comparing both regimes in terms of accuracy (cf. \Cref{fig:resnet_energy_scaling_vds}B), the small-scale regime performs better for smaller amounts of data than the large-scale regime, while for growing training dataset size, the large-scale regime provides a steeper curve, i.e., substantial improvement of the prediction accuracy. 
Keeping the GBS constant while increasing dataset size will, in theory, produce counteracting effects on training time efficiency:
Increasing the dataset size increases the number of gradient updates and thus, overall training time, while decreasing the samples per gradient update (LBS) accelerates each individual gradient update due to smaller computational workloads.
This effect can be observed in \Cref{fig:resnet_energy_scaling_vds}A.
The main distinction between the two regimes lies in the higher number of gradient updates in the small-scale regime (GBS=256), associated with a considerable workload, whereas in the large-scale regime (GBS=8\,192) this workload is generally small.
In the small-scale regime, the two opposing effects initially compensate each other, yielding constant training times, but as the number of GPUs increases, training time becomes dominated by the growing number of gradient updates.
In the large-scale regime, the training time stays constant throughout, as the two effects keep counteracting each other. 
In both scaling experiments, the energy consumption scales almost linearly with the number of samples (cf. \Cref{fig:resnet_energy_scaling_vds}C).
The energy per sample slightly increases during the scaling experiments from \SI{0.0221}{\watt\hour} to \SI{0.0260}{\watt\hour} (small-scale) and from \SI{0.0238}{\watt\hour} to \SI{0.0260}{\watt\hour} (large-scale) per sample (neglecting $\mathrm{n < 4}$ GPUs).
This also corresponds to an approximately linear scaling of energy consumption with respect to GPU hours (cf. \Cref{fig:resnet_energy_scaling_vds}D).

Our results show that achieving higher accuracy through larger training datasets requires approximately proportionally more energy.

\subsection{FourCastNet}
\label{sec:fcn}

\subsubsection{Setup}
We used the original FourCastNet model~\cite{pathak2022fourcastnet}, for which trainable code was published~\cite{fcn_code}, and trained it on the ERA5 data~\cite{hersbach2020era5}, using data between 1979 and 2015 for training, 2016 and 2017 for validation, and 2019 as the test set.
Following the original training scheme, we applied pre-training, i.e., predicting one \SI{6}{\hour} time frame given the previous one, for 80 epochs, and fine-tuning, i.e., two consecutive $\mathrm{6 \, h}$ time steps for a \SI{12}{\hour} prediction using autoregressive rollout, for 50 epochs.
We used the \texttt{CosineAnnealingLR} learning-rate scheduler and the \texttt{Adam} optimizer for gradient updates with learning rates of \num{0.0005} for pre-training and \num{0.0001} for fine-tuning.
Accuracy was evaluated on the commonly used Z500 RMSE, i.e., geopotential height at \SI{500}{\hecto\pascal}, for the \SI{6}{\hour} (pre-training) and the \SI{12}{\hour} (fine-tuning) prediction using the mean of 36 samples of the test set.
Due to the high computational resource demands (GPU hours and energy) of training, we generally forwent running multiple experiments per setup. 
To estimate statistical fluctuations, we evaluated training time and energy for a single setup of parallelization parameters (GBS=64, LBS=1, 64=GPUs, 20 epochs) using three runs, which yielded an average training time of $\SI{4}{\hour} \, \SI{23}{\minute} \pm \SI{9}{\minute}$,  corresponding to $\num{280} \pm \num{10} \, \mathrm{GPUh}$, and an energy consumption of $\num{52} \pm \SI{2}{\kilo\watt\hour}$.
Given the enormous size of the individual samples, the LBS is limited to four samples on the 40GB-A100 GPU.
We further utilized the H100 GPUs in these experiments, which allow higher LBS due to their larger memory capacity.
The energies reported for A100 are the sum of RAM, CPU, and GPU energies.
For H100 nodes, measuring the RAM energy is not supported, but its contribution is generally negligible.

\subsubsection{Training on a Constant Dataset}
Similar to the first set of experiments for ResNet (~\cref{sec:resnet_cDS}), we increased the number of A100 GPUs for training on the full dataset with a constant GBS or a constant LBS.
For the pre-training with a constant LBS=1, we increased the number of GPUs from $n=\mathrm{4}$ to $n=\mathrm{256}$, resulting in the same numbers for GBSs.
For fine-tuning with LBS=1, using the weights of the corresponding pre-training, we studied $n=\mathrm{16}$ to $n=\mathrm{256}$ GPUs.
\begin{figure*}
  \centering
  \includegraphics[width=\linewidth]{ 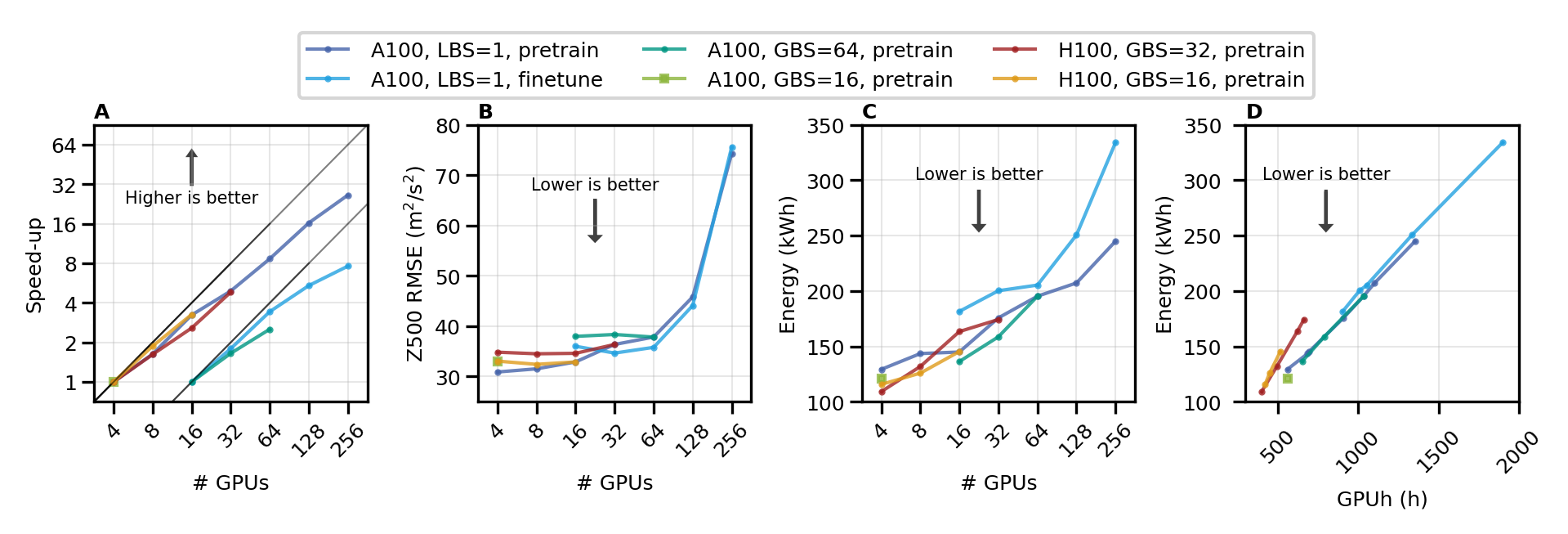}
  \caption{Scaling experiments for FourCastNet trained on the ERA5 dataset.
  A constant local batch size (LBS) or constant global batch size (GBS) is used for pre-training as well as fine-tuning FourCastNet.
  Both H100 and A100 GPUs were utilized.
  Speed-up (A), top-1 error (B), and energy consumption (C) depending on the number of GPUs are shown.
  Additionally, the energy consumption versus GPU hours (D) are depicted.}
  \Description{}
  \label{fig:fcn_scaling_general}
\end{figure*}
Already after quadrupling the number of GPUs, the speed-up clearly diverges from ideal scaling (cf. \Cref{fig:fcn_energy_scaling}A). 
Large batch effects also occur much earlier for comparably low GBS (cf. \Cref{fig:fcn_energy_scaling}B): 
The Z500 RMSE already increases for the pre-training slightly beyond GBS $\mathrm{> 4}$ and becomes steeper after GBS $\mathrm{> 64}$.
For the fine-tuning, pronounced large batch effects are observed for GBS $\mathrm{> 64}$.
Increasing the number of GPUs also increases the energy consumption.
While the consumed energy for pre-training increases almost linearly when doubling the number of GPUs, the energy required for fine-tuning increases exponentially (cf. \Cref{fig:fcn_energy_scaling}C).
Additionally, fine-tuning generally requires more energy, since the autoregressive scheme leads to longer training times per epoch for the same number of GPUs utilized.
The energy per GPUh increases for both trainings with a similar linear factor (cf. \Cref{fig:fcn_energy_scaling}D).

Scaling experiments for constant GBS using only pre-training were performed with three setups: GBS=64 on the A100 GPUs and GBS=16 and GBS=32 on the H100 GPUs.
Again, a sub-linear speed-up is obtained, while the Z500 RMSE stays approximately constant (cf. \Cref{fig:fcn_energy_scaling}A,B).
The energy increases linearly with the amount of GPU hours; however, differences between H100 and A100 nodes can be observed, with the gradient for H100 nodes being steeper (cf. \Cref{fig:fcn_energy_scaling}C,D).
In \Cref{tbl:fcn_h100_vs_a100}, we compare all three FourCastNet pre-training experiments for identical GBS, LBS, and GPU counts between the different accelerator types.
Training times and GPUh on H100 GPUs are reduced by \SI{25}{\percent} to \SI{30}{\percent} compared to A100 GPUs.
However, the energy consumed differs by less than \SI{5}{\percent}, due to the higher power consumption of the H100 nodes. 
This is discussed further in \Cref{sec:power_scaling} and \Cref{sec:power_profiles}.
\begin{table}
    \begin{center}
        \small
        \caption{Experiments for pre-training FourCastNet on ERA5 using H100 and A100 GPUs with specific number of GPUs, local batch size (LBS), and global batch size (GBS).
        Energy consumptions, energy consumptions per node, training times, GPU hours, and the Z500 RMSEs are listed.}
        \begin{tabular}{lllllll}
            \toprule
            Device & GPUs & LBS & GBS & Energy & Runtime & GPUh \\ 
              &   &  &  & (kWh) & (h) & (h) \\ 
            \midrule 
            H100 & 4 & 4 & 16 & 115.63 & 105.32 & 421.26\\ 
            A100 & 4 & 4 & 16 & 120.70 & 139.92 & 559.70\\ 
            H100 & 16 & 1 & 16 & 145.31  & 32.26 & 516.18\\ 
            A100 & 16 & 1 & 16 & 144.87 & 43.28 & 692.55\\ 
            H100 & 32 & 1 & 32 & 174.25 & 20.72 & 663.13\\ 
            A100 & 32 & 1 & 32 & 175.72 & 28.42 & 909.31\\ 
            \bottomrule
        \end{tabular}
        \label{tbl:fcn_h100_vs_a100}
    \end{center}
\end{table}
\newline
Regarding energy consumption and accuracy, training FourCastNet with a low number of GPUs is generally more beneficial, as speed-up is highly inefficient.
While scaling the number of GPUs reduces training time considerably, it also yields an enormous increase in resource consumption (GPUh). 
For example, increasing from 16 to 64 GPUs for a constant GBS=64 increases the GPUh from \SI{651}{\hour} to \SI{1034}{\hour} and the energy consumption from \SI{136}{\kilo\watt\hour} to \SI{195}{\kilo\watt\hour}, while the training time is reduced from \SI{41}{\hour} to \SI{16}{\hour}.

\subsection{GPU Power Profile}
\label{sec:power_profiles}

\begin{figure}
  \centering
  \includegraphics[width=\linewidth]{ 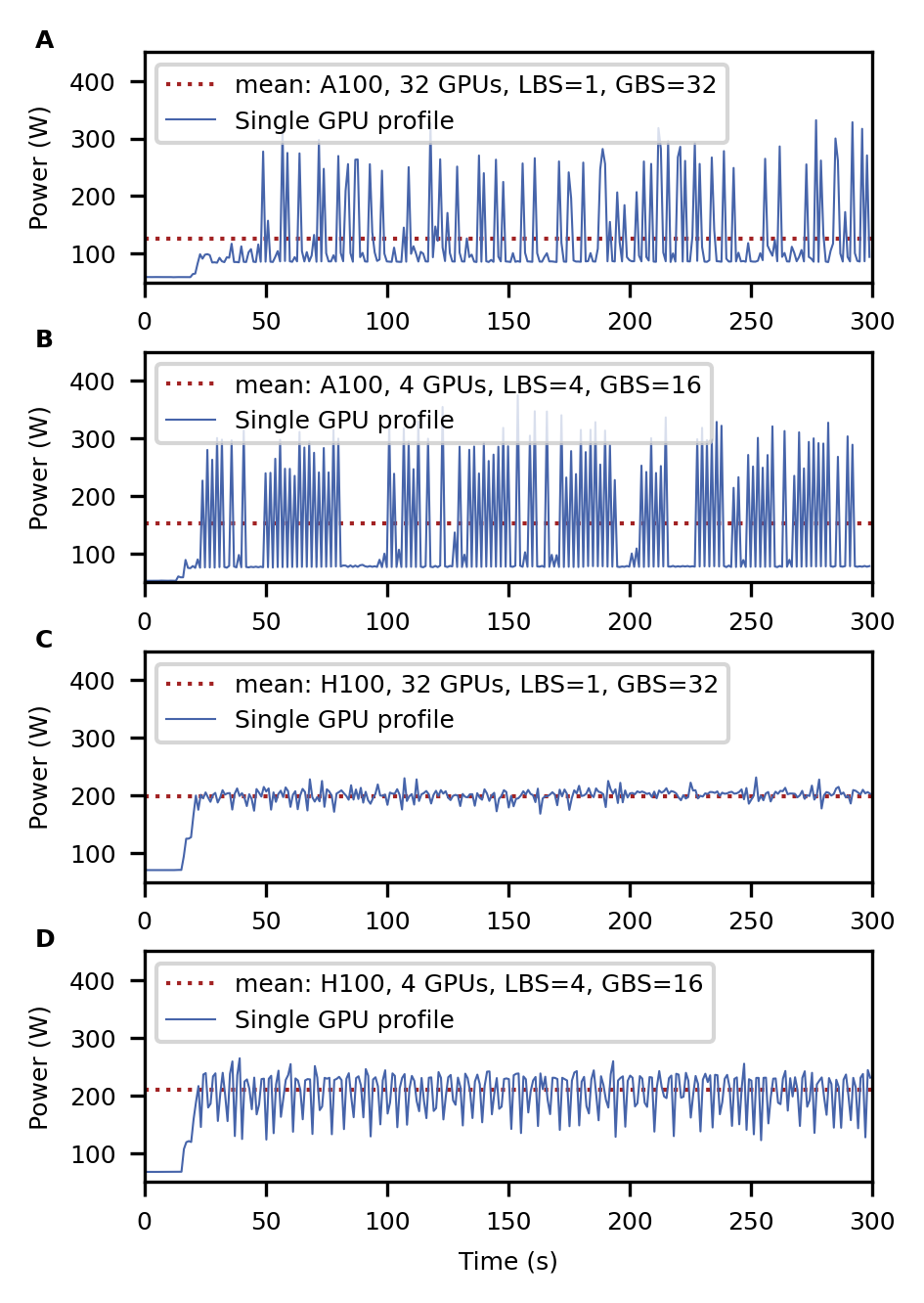}
  \caption{GPU power profiles of experiments for FourCastNet pre-trained on ERA5 conducted on A100 (A, B) and H100 (C, D) GPUs with corresponding local batch sizes (LBS), global batch sizes (GBS), and numbers of GPUs as indicated in the legend.
  Profiles are depicted for arbitrarily selected single-GPU devices used for the experiments.
  Additionally, the mean over time and all GPUs for the corresponding experiment is shown.}
  \Description{}
  \label{fig:fcn_power_profiles}
\end{figure}
Running training on H100 and A100 GPUs has a significant impact on energy consumption.
In \Cref{fig:fcn_power_profiles}, we show the first $\mathrm{300 \, s}$ of two power profiles for each A100 and H100 experiments pre-training FourCastNet.
The profiles for the A100 GPUs have broad fluctuations and fall regularly back to a baseline. 
Similar profiles are observed for training ResNet on A100 GPUs.
Beyond indicating inefficient utilization, such GPU power fluctuations can also accelerate hardware degradation due to thermal stress and thus shorten GPU lifespan~\cite{ostrouchov2020gpu}.
The profiles for the H100 GPUs are narrower than for the A100 GPUs and do not fall back to a baseline.
Hence, these experiments indicate that the H100 GPUs are running more efficiently.
Lower LBS, i.e., a lower memory utilization, further narrows the power profile. 
We hypothesize that data loading plays a major factor in these differences.

\subsection{Power Draw Scaling}
\label{sec:power_scaling}

\begin{figure*}
  \centering
  \includegraphics[width=\linewidth]{ 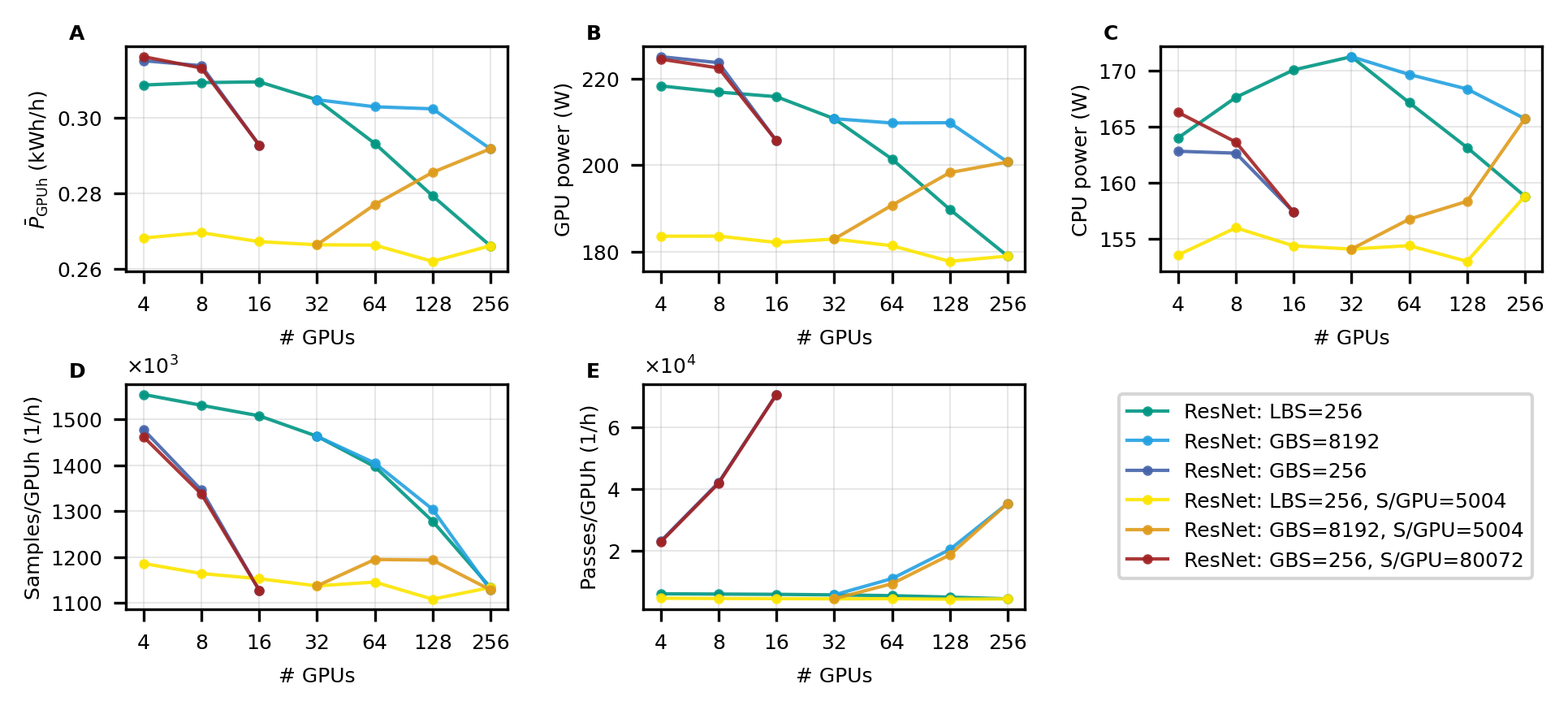}
  \caption{Energy per GPUh $\bar{P}_{\mathrm{GPUh}}$ (A), single mean GPU (B) and CPU socket (C) power, samples per GPU and hour (D), and gradient updates (forward/backward passes) per GPU and hour (E) for all conducted ResNet experiments.
  These experiments ran with a constant local batch size (LBS) or a constant global batch size (GBS) using a fixed overall dataset of \num{1281167} samples (used, unless otherwise specified) or a fixed number of samples per GPU (S/GPU).}
  \Description{}
  \label{fig:resnet_energy_scaling}
\end{figure*}
\begin{figure*}
  \centering
  \includegraphics[width=\linewidth]{ 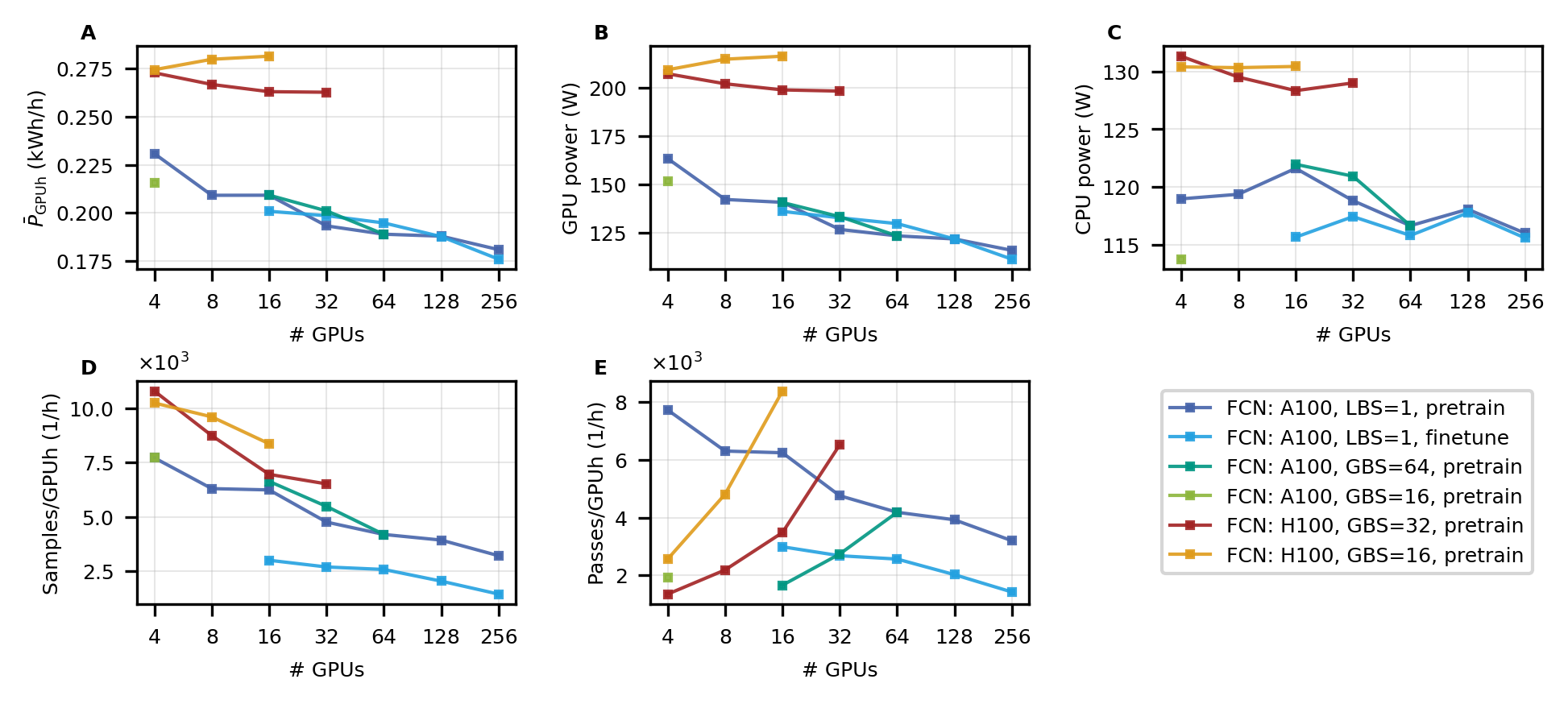}
  \caption{Energy per GPUh $\bar{P}_{\mathrm{GPUh}}$ (A), single mean GPU (B) and CPU socket (C) power, samples per GPU and hour (D), and gradient updates (forward/backward passes) per GPU and hour (E) for all conducted FourCastNet experiments.
  These experiments ran on A100, as well as H100 GPUs, using a constant local batch size (LBS) or a constant global batch size (GBS).
  pre-training as well as fine-tuning for FourCastNet at the ERA5 data was performed.}
  \Description{}
  \label{fig:fcn_energy_scaling}
\end{figure*}

\Cref{sec:resnet_cDS} to \Cref{sec:fcn} investigated training time, accuracy, energy consumption, and the relation between energy and GPU hours for each scaling experiment. 
We revisit these experiments to explain how energy consumption is influenced by parallelization parameters (LBS, GBS. GPU count).
In \Cref{fig:resnet_energy_scaling} (ResNet) and \Cref{fig:fcn_energy_scaling} (FourCastNet), we show different metrics related to the energy consumption for all experiments.
An obvious conclusion is that the consumed energy scales linearly with the amount of GPU hours, as already noted.
Thus, one might expect the energy per GPU hour $\bar{P}_{\mathrm{GPUh}}$ to remain constant (cf. \Cref{fig:resnet_energy_scaling}A and \Cref{fig:fcn_energy_scaling}A), thereby providing an appropriate metric to discuss energy efficiency, which allows a direct mapping from resources (GPUh) and training time to the energy consumption.
Indeed, it stays within intervals of (\SI{0.18}{\kilo\watt}, \SI{0.24}{\kilo\watt}) (FourCastNet, A100), (\SI{0.26}{\kilo\watt}, \SI{0.28}{\kilo\watt}) (FourCastNet, H100), and (\SI{0.26}{\kilo\watt}, \SI{0.32}{\kilo\watt}) (ResNet, A100), however, these intervals differ both in position (magnitude of power draw) and width.
$\bar{P}_{\mathrm{GPUh}}$ equals the sum of the mean (averaged across all corresponding devices and time) power draws of all devices (CPU, RAM, GPU). 
The power draw shows stronger variations for GPUs than for CPUs, while the RAM power draw is negligible with a energy contribution of $\mathrm{< 4 \, \%}$ to the total energy for experiments with at least four GPUs (cf. \Cref{fig:resnet_energy_scaling}B, C  \Cref{fig:fcn_energy_scaling}B, C).
Thus, energy efficiency in essence depends on GPU power draw (and CPU, to a smaller extent), as trivially expected.

Unraveling the GPU power draw with respect to the parallelization parameters and the steps within the training workflow manifests as the key challenge.
Towards this end, we discuss deviations in GPU power draw for the different scaling experiments in more detail, setting it into relation with the workload per time, i.e., number of samples and the number of gradient updates (forward/backward passes) per GPU hour (\Cref{fig:resnet_energy_scaling}D, E and \Cref{fig:fcn_energy_scaling}D, E).
\newline
For both FourCastNet (\Cref{fig:fcn_energy_scaling}) and ResNet (\Cref{fig:resnet_energy_scaling}) experiments with a fixed-size dataset and constant LBS, the number of samples and forward--backward passes processed per GPU hour decreases with increasing number of GPUs, due to non-ideal scaling.
This decay is reflected by a decrease in GPU power draw.
The experiments with a constant dataset and constant GBS show opposing behavior between samples and passes per GPUh.
The number of samples processed per GPUh is decreasing, while the number of forward--backward passes conducted per GPUh is increasing with higher number of GPUs, due to non-ideal speedup and thus, less throughput per gradient update.
For ResNet (\Cref{fig:fcn_energy_scaling}), the result is a decrease in GPU power draw, i.e., sample throughput has a greater impact on the power draw than the number of passes.
For FourCastNet (\Cref{fig:resnet_energy_scaling}), the GPU power draw for the A100 experiments with GBS=64 and the H100 experiments with GBS=32 is decreasing, while the GPU power draw for the H100 experiments with GBS=16 is increasing. 
Comparing the two H100 scaling experiments reveals a turnover point at which the influence of passes and samples per GPUh to GPU power draw shifts.

For the ResNet experiment (\Cref{fig:fcn_energy_scaling}) with the dataset size scaled proportionally to GPU count, i.e., keeping samples per GPU (S/GPU) constant, and with constant LBS=256, all GPUs are processing the same number of samples and passes per epoch during scaling the GPUs.
This is reflected by constant samples and passes per GPUh, resulting in constant GPU power.
The two ResNet experiments with scaled dataset sizes and constant GBS show different trends.
For both, the passes per GPUh are increasing, but the samples per GPUh are approximately constant for GBS=256 and decreasing for GBS=8\,192.
These differences can be attributed to the substantially higher number of passes for the GBS=256 experiments compared to the GBS=8\,192 experiments, which decelerates the training (cf. \Cref{sec:resnet_cDS}A) and, therefore, reduces the sample throughput.
The GPU power for the GBS=256 experiments is decreasing, and for the GBS=8\,192 experiments increasing.
Considering the trends of the GPU power, it is rather dominated by the samples per GPUh.

Generally, the GPU power during data-parallel DL, seems to be related to the samples per GPUh and passes per GPUh.
A high throughput in terms of number of samples per GPUh, leads to higher GPU power.

\section{Conclusion}
We performed scaling experiments for ResNet50 trained on ImageNet-2012 and FourCastNet trained on ERA5 by studying the influence of parallelization parameters, i.e., the number of GPUs, global batch size (GBS), and local batch size (LBS), on the training time, accuracy, and energy consumption.

With a constant dataset, ResNet scales nearly linearly, while FourCastNet significantly deviates from linear speed-up after quadrupling GPUs, making it inefficient in terms of GPU hours.
Both models suffer from large batch effects, but they emerge earlier for FourCastNet.
The accuracy drops beyond GBS=8\,192 for ResNet and beyond GBS=32 (pre-training) and GBS=64 (fine-tuning) for FourCastNet.
For ResNet, increasing the dataset size with the GPU count keeps the training time mostly constant.
However, for scaling with a constant GBS, the increased number of gradient updates for high LBS leads to slightly growing training times. 
When scaling the dataset size in ResNet with a constant LBS=256, opposing effects on accuracy are observed: increasing dataset size improves accuracy, while the large-batch effect tends to diminish it.

We found that the energy consumption linearly scales with the number of GPU hours.
The corresponding factors are approximately \SI{0.27}{\kilo\watt\hour\per\hour} to \SI{0.31}{\kilo\watt\hour\per\hour} for ResNet, \SI{0.18}{\kilo\watt\hour\per\hour} to \SI{0.23}{\kilo\watt\hour\per\hour} for FourCastNet on A100 GPUs, and \SI{0.26}{\kilo\watt\hour\per\hour} to \SI{0.28}{\kilo\watt\hour\per\hour} for FourCastNet on H100 GPUs.
However, these are rough estimates, since the GPUs run at different power depending on the model and parallelization parameters.
In particular, we found that the samples per GPUh and gradient updates per GPUh systematically influence the GPU power.
Deviations of CPU power are smaller, and the contribution of the RAM power to the energy is generally negligible.
For training on A100 GPUs, high fluctuations in the power profile and a rather low mean power for higher numbers of GPUs were observed.
This utilization is not sustainable, since power fluctuations in processing units are known to decrease their lifespan.
H100 GPUs run with less and narrower fluctuations at higher power draws.

Our study demonstrates how parallelization affects energy consumption and GPU power draw. 
Understanding such relations is crucial for enabling resource-efficient data-parallel deep learning, where training time, predictive performance, and energy consumption have to be balanced.

\section*{Acknowledgment}
This work is supported by the German Federal Ministry of Research, Technology and Space (BMFTR) and the German federal states as part of the National High-Performance Computing (NHR) joint funding program (http://www.nhr-verein.de/en/our-partners), the BMFTR 01LK2313A SMARTWEATHER21-SCC-2 grant, and by the Helmholtz AI platform grant.
The authors gratefully acknowledge the computing time made available to them on the high-performance computer HoreKa at the NHR Center KIT via the SmartWeather21-p0021348 NHR large project.

\section*{References}
\bibliographystyle{IEEEtran_mod}
\bibliography{software}

\begin{thebibliography}{10}
\providecommand{\url}[1]{#1}
\csname url@samestyle\endcsname
\providecommand{\newblock}{\relax}
\providecommand{\bibinfo}[2]{#2}
\providecommand{\BIBentrySTDinterwordspacing}{\spaceskip=0pt\relax}
\providecommand{\BIBentryALTinterwordstretchfactor}{4}
\providecommand{\BIBentryALTinterwordspacing}{\spaceskip=\fontdimen2\font plus
\BIBentryALTinterwordstretchfactor\fontdimen3\font minus \fontdimen4\font\relax}
\providecommand{\BIBforeignlanguage}[2]{{%
\expandafter\ifx\csname l@#1\endcsname\relax
\typeout{** WARNING: IEEEtran.bst: No hyphenation pattern has been}%
\typeout{** loaded for the language `#1'. Using the pattern for}%
\typeout{** the default language instead.}%
\else
\language=\csname l@#1\endcsname
\fi
#2}}
\providecommand{\BIBdecl}{\relax}
\BIBdecl

\bibitem{ben2019demystifying}
T.~Ben-Nun and T.~Hoefler, ``Demystifying parallel and distributed deep learning: An in-depth concurrency analysis,'' \emph{ACM Comput. Surv.}, vol.~52, no.~4, 8 2019, \href{https://doi.org/10.1145/3320060}{doi:10.1145/3320060}.

\bibitem{keuper2016distributed}
J.~Keuper and F.-J. Preundt, ``Distributed training of deep neural networks: Theoretical and practical limits of parallel scalability,'' \emph{2016 2nd Workshop on Machine Learning in HPC Environments (MLHPC)}, pp. 19--26, 2016, \href{https://doi.org/10.1109/MLHPC.2016.006}{doi:10.1109/MLHPC.2016.006}.

\bibitem{li2020pytorchdistributed}
S.~Li, Y.~Zhao, R.~Varma, O.~Salpekar, P.~Noordhuis, T.~Li, A.~Paszke, J.~Smith, B.~Vaughan, P.~Damania, and S.~Chintala, ``Pytorch distributed: Experiences on accelerating data parallel training,'' \emph{arXiv preprint arXiv:2006.15704}, 2020, \href{https://doi.org/10.48550/arXiv.2006.15704}{doi:10.48550/arXiv.2006.15704}.

\bibitem{goyal2017accurate}
P.~Goyal, P.~Doll{\'a}r, R.~Girshick, P.~Noordhuis, L.~Wesolowski, A.~Kyrola, A.~Tulloch, Y.~Jia, and K.~He, ``Accurate, large minibatch sgd: Training imagenet in 1 hour,'' \emph{arXiv preprint arXiv:1706.02677}, 2017, \href{https://doi.org/10.48550/arXiv.1706.02677}{doi:10.48550/arXiv.1706.02677}.

\bibitem{you2017large}
Y.~You, I.~Gitman, and B.~Ginsburg, ``Large batch training of convolutional networks,'' \emph{arXiv preprint arXiv:1708.03888}, 2017, \href{https://doi.org/10.48550/arXiv.1708.03888}{doi:10.48550/arXiv.1708.03888}.

\bibitem{yamazaki2019yet}
M.~Yamazaki, A.~Kasagi, A.~Tabuchi, T.~Honda, M.~Miwa, N.~Fukumoto, T.~Tabaru, A.~Ike, and K.~Nakashima, ``Yet another accelerated sgd: Resnet-50 training on imagenet in 74.7 seconds,'' \emph{arXiv preprint arXiv:1903.12650}, 2019, \href{https://doi.org/10.48550/arXiv.1903.12650}{doi:10.48550/arXiv.1903.12650}.

\bibitem{malladi2022sdes}
\BIBentryALTinterwordspacing
S.~Malladi, K.~Lyu, A.~Panigrahi, and S.~Arora, ``On the sdes and scaling rules for adaptive gradient algorithms,'' \emph{Advances in Neural Information Processing Systems}, vol.~35, pp. 7697--7711, 2022. Online \href{https://proceedings.neurips.cc/paper_files/paper/2022/file/32ac710102f0620d0f28d5d05a44fe08-Paper-Conference.pdf}{available}
\BIBentrySTDinterwordspacing

\bibitem{debus2023reporting}
C.~Debus, M.~Piraud, A.~Streit, F.~Theis, and M.~G{\"o}tz, ``Reporting electricity consumption is essential for sustainable ai,'' \emph{Nat. Mach. Intell.}, vol.~5, no.~11, pp. 1176--1178, 11 2023, \href{https://doi.org/10.1038/s42256-023-00750-1}{doi:10.1038/s42256-023-00750-1}.

\bibitem{dauner2025energy}
M.~Dauner and G.~Socher, ``Energy costs of communicating with ai,'' \emph{Front. Commun.}, vol.~10, 6 2025, \href{https://doi.org/10.3389/fcomm.2025.1572947}{doi:10.3389/fcomm.2025.1572947}.

\bibitem{he2016deep}
K.~He, X.~Zhang, S.~Ren, and J.~Sun, ``Deep residual learning for image recognition,'' \emph{Proceedings of the IEEE conference on computer vision and pattern recognition}, pp. 770--778, 2016, \href{https://doi.org/10.1109/CVPR.2016.90}{doi:10.1109/CVPR.2016.90}.

\bibitem{deng2009imagenet}
J.~Deng, W.~Dong, R.~Socher, L.-J. Li, K.~Li, and L.~Fei-Fei, ``Imagenet: A large-scale hierarchical image database,'' \emph{2009 IEEE conference on computer vision and pattern recognition}, pp. 248--255, 2009, \href{https://doi.org/10.1109/CVPR.2009.5206848}{doi:10.1109/CVPR.2009.5206848}.

\bibitem{pathak2022fourcastnet}
J.~Pathak, S.~Subramanian, P.~Harrington, S.~Raja, A.~Chattopadhyay, M.~Mardani, T.~Kurth, D.~Hall, Z.~Li, K.~Azizzadenesheli \emph{et~al.}, ``Fourcastnet: A global data-driven high-resolution weather model using adaptive fourier neural operators,'' \emph{arXiv preprint arXiv:2202.11214}, 2022, \href{https://doi.org/10.48550/arXiv.2202.11214}{doi:10.48550/arXiv.2202.11214}.

\bibitem{fcn_code}
\BIBentryALTinterwordspacing
``Fourcastnet,'' NVIDIA Corporation, 2022. Online \href{https://github.com/NVlabs/FourCastNet}{available}
\BIBentrySTDinterwordspacing

\bibitem{hersbach2020era5}
H.~Hersbach, B.~Bell, P.~Berrisford, S.~Hirahara, A.~Hor{\'a}nyi, J.~Mu{\~n}oz-Sabater, J.~Nicolas, C.~Peubey, R.~Radu, D.~Schepers \emph{et~al.}, ``The era5 global reanalysis,'' \emph{Q. J. R. Meteorol. Soc}, vol. 146, no. 730, pp. 1999--2049, 2020, \href{https://doi.org/10.1002/qj.3803}{doi:10.1002/qj.3803}.

\bibitem{schwartz2020green}
R.~Schwartz, J.~Dodge, N.~A. Smith, and O.~Etzioni, ``Green ai,'' \emph{CACM}, vol.~63, no.~12, pp. 54--63, 11 2020, \href{https://doi.org/10.1145/3381831}{doi:10.1145/3381831}.

\bibitem{luccioni2019quantifying}
\BIBentryALTinterwordspacing
S.~Luccioni, V.~Schmidt, A.~Lacoste, and T.~Dandres, ``Quantifying the carbon emissions of machine learning,'' \emph{NeurIPS 2019 Workshop on Tackling Climate Change with Machine Learning}, 2019. Online \href{https://www.climatechange.ai/papers/neurips2019/22}{available}
\BIBentrySTDinterwordspacing

\bibitem{mehlin2023towards}
V.~Mehlin, S.~Schacht, and C.~Lanquillon, ``Towards energy-efficient deep learning: An overview of energy-efficient approaches along the deep learning lifecycle,'' \emph{arXiv preprint arXiv:2303.01980}, 2023, \href{https://doi.org/10.48550/arXiv.2303.01980}{doi:10.48550/arXiv.2303.01980}.

\bibitem{xu2023energy}
Y.~Xu, S.~Mart{\'\i}nez-Fern{\'a}ndez, M.~Martinez, and X.~Franch, ``Energy efficiency of training neural network architectures: an empirical study,'' \emph{arXiv preprint arXiv:2302.00967}, 2023, \href{https://doi.org/10.48550/arXiv.2302.00967}{doi:10.48550/arXiv.2302.00967}.

\bibitem{tripp2024measuring}
C.~E. Tripp, J.~Perr-Sauer, J.~Gafur, A.~Nag, A.~Purkayastha, S.~Zisman, and E.~A. Bensen, ``Measuring the energy consumption and efficiency of deep neural networks: An empirical analysis and design recommendations,'' \emph{arXiv preprint arXiv:2403.08151}, 2024, \href{https://doi.org/10.48550/arXiv.2403.08151}{doi:10.48550/arXiv.2403.08151}.

\bibitem{you2023zeus}
\BIBentryALTinterwordspacing
J.~You, J.-W. Chung, and M.~Chowdhury, ``Zeus: Understanding and optimizing {GPU} energy consumption of {DNN} training,'' \emph{20th USENIX Symposium on Networked Systems Design and Implementation (NSDI 23)}, pp. 119--139, 2023. Online \href{https://www.usenix.org/conference/nsdi23/presentation/you}{available}
\BIBentrySTDinterwordspacing

\bibitem{caspart2022precise}
R.~Caspart, S.~Ziegler, A.~Weyrauch, H.~Obermaier, S.~Raffeiner, L.~P. Schuhmacher, J.~Scholtyssek, D.~Trofimova, M.~Nolden, I.~Reinartz \emph{et~al.}, ``Precise energy consumption measurements of heterogeneous artificial intelligence workloads,'' \emph{International Conference on High Performance Computing}, pp. 108--121, 2022, \href{https://doi.org/10.1007/978-3-031-23220-6\_8}{doi:10.1007/978-3-031-23220-6\_8}.

\bibitem{yarally2023uncovering}
T.~Yarally, L.~Cruz, D.~Feitosa, J.~Sallou, and A.~Van~Deursen, ``Uncovering energy-efficient practices in deep learning training: Preliminary steps towards green ai,'' \emph{2023 IEEE/ACM 2nd International Conference on AI Engineering--Software Engineering for AI (CAIN)}, pp. 25--36, 2023, \href{https://doi.org/10.1109/CAIN58948.2023.00012}{doi:10.1109/CAIN58948.2023.00012}.

\bibitem{geissler2024spend}
D.~Geissler, B.~Zhou, S.~Suh, and P.~Lukowicz, ``Spend more to save more (sm2): An energy-aware implementation of successive halving for sustainable hyperparameter optimization,'' 2024.

\bibitem{geissler2024power}
D.~Gei{\ss}ler, B.~Zhou, M.~Liu, S.~Suh, and P.~Lukowicz, ``The power of training: How different neural network setups influence the energy demand,'' \emph{International Conference on Architecture of Computing Systems}, pp. 33--47, 2024, \href{https://doi.org/10.1007/978-3-031-66146-4\_3}{doi:10.1007/978-3-031-66146-4\_3}.

\bibitem{frey2022benchmarking}
N.~C. Frey, B.~Li, J.~McDonald, D.~Zhao, M.~Jones, D.~Bestor, D.~Tiwari, V.~Gadepally, and S.~Samsi, ``Benchmarking resource usage for efficient distributed deep learning,'' \emph{2022 IEEE High Performance Extreme Computing Conference (HPEC)}, pp. 1--8, 2022, \href{https://doi.org/10.1109/HPEC55821.2022.9926375}{doi:10.1109/HPEC55821.2022.9926375}.

\bibitem{koszczal2023performance}
G.~Koszcza{\l}, J.~Dobrosolski, M.~Matuszek, and P.~Czarnul, ``Performance and energy aware training of a deep neural network in a multi-gpu environment with power capping,'' \emph{European Conference on Parallel Processing}, pp. 5--16, 2023, \href{https://doi.org/10.1007/978-3-031-48803-0\_1}{doi:10.1007/978-3-031-48803-0\_1}.

\bibitem{liang2024communication}
F.~Liang, Z.~Zhang, H.~Lu, V.~Leung, Y.~Guo, and X.~Hu, ``Communication-efficient large-scale distributed deep learning: A comprehensive survey,'' \emph{arXiv preprint arXiv:2404.06114}, 2024, \href{https://doi.org/10.48550/arXiv.2404.06114}{doi:10.48550/arXiv.2404.06114}.

\bibitem{shen2024efficient}
L.~Shen, Y.~Sun, Z.~Yu, L.~Ding, X.~Tian, and D.~Tao, ``On efficient training of large-scale deep learning models,'' \emph{ACM Comput. Surv.}, vol.~57, no.~3, 11 2024, \href{https://doi.org/10.1145/3700439}{doi:10.1145/3700439}.

\bibitem{vogels_powersgd_2019}
\BIBentryALTinterwordspacing
T.~Vogels, S.~P. Karimireddy, and M.~Jaggi, ``{PowerSGD}: {Practical} {Low}-{Rank} {Gradient} {Compression} for {Distributed} {Optimization},'' \emph{Advances in {Neural} {Information} {Processing} {Systems}}, vol.~32, 2019. Online \href{https://proceedings.neurips.cc/paper_files/paper/2019/file/d9fbed9da256e344c1fa46bb46c34c5f-Paper.pdf}{available}
\BIBentrySTDinterwordspacing

\bibitem{abrahamyan2022compression}
L.~Abrahamyan, Y.~Chen, G.~Bekoulis, and N.~Deligiannis, ``{Learned Gradient Compression for Distributed Deep Learning},'' \emph{IEEE Trans. Neural Netw. Learn}, vol.~33, no.~12, pp. 7330--7344, 12 2022, \href{https://doi.org/10.1109/TNNLS.2021.3084806}{doi:10.1109/TNNLS.2021.3084806}.

\bibitem{niu_hogwild_2011}
F.~Niu, B.~Recht, C.~Re, and S.~J. Wright, ``{HOGWILD}!: {A} {Lock}-{Free} {Approach} to {Parallelizing} {Stochastic} {Gradient} {Descent},'' \emph{arXiv preprint arXiv:1106.5730}, 11 2011, \href{https://doi.org/10.48550/arXiv.1106.5730}{doi:10.48550/arXiv.1106.5730}.

\bibitem{lin_dont_2020}
T.~Lin, S.~U. Stich, K.~K. Patel, and M.~Jaggi, ``Don't {Use} {Large} {Mini}-{Batches}, {Use} {Local} {SGD},'' \emph{arXiv preprint arXiv:1808.07217}, 2 2020, \href{https://doi.org/10.48550/arXiv.1808.07217}{doi:10.48550/arXiv.1808.07217}.

\bibitem{coquelin2022accelerating}
D.~Coquelin, C.~Debus, M.~G{\"o}tz, F.~von~der Lehr, J.~Kahn, M.~Siggel, and A.~Streit, ``Accelerating neural network training with distributed asynchronous and selective optimization (daso),'' \emph{J. Big Data}, vol.~9, no.~1, p.~14, 2 2022, \href{https://doi.org/10.1186/s40537-021-00556-1}{doi:10.1186/s40537-021-00556-1}.

\bibitem{ahn2024efficient}
S.~Ahn, S.~Lee, H.~Choi, and J.~Lee, ``Efficient data-parallel distributed dnn training for big dataset under heterogeneous gpu cluster,'' \emph{2024 IEEE International Conference on Big Data (BigData)}, pp. 179--188, 2024, \href{https://doi.org/10.1145/567752.567774}{doi:10.1145/567752.567774}.

\bibitem{chen2021empirical}
X.~Chen, S.~Xie, and K.~He, ``An empirical study of training self-supervised vision transformers,'' \emph{Proceedings of the IEEE/CVF international conference on computer vision}, pp. 9640--9649, 2021, \href{https://doi.org/10.1109/ICCV48922.2021.00950}{doi:10.1109/ICCV48922.2021.00950}.

\bibitem{kurth2023fourcastnet}
T.~Kurth, S.~Subramanian, P.~Harrington, J.~Pathak, M.~Mardani, D.~Hall, A.~Miele, K.~Kashinath, and A.~Anandkumar, ``Fourcastnet: Accelerating global high-resolution weather forecasting using adaptive fourier neural operators,'' \emph{Proceedings of the platform for advanced scientific computing conference}, pp. 1--11, 2023, \href{https://doi.org/10.1145/3592979.3593412}{doi:10.1145/3592979.3593412}.

\bibitem{gutierrez2023perun}
J.~P. Guti{\'e}rrez Hermosillo~Muriedas, K.~Fl{\"u}gel, C.~Debus, H.~Obermaier, A.~Streit, and M.~G{\"o}tz, ``Perun: Benchmarking energy consumption of high-performance computing applications,'' \emph{European Conference on Parallel Processing}, pp. 17--31, 2023, \href{https://doi.org/10.1007/978-3-031-39698-4\_2}{doi:10.1007/978-3-031-39698-4\_2}.

\bibitem{ostrouchov2020gpu}
G.~Ostrouchov, D.~Maxwell, R.~A. Ashraf, C.~Engelmann, M.~Shankar, and J.~H. Rogers, ``Gpu lifetimes on titan supercomputer: Survival analysis and reliability,'' \emph{SC20: International Conference for High Performance Computing, Networking, Storage and Analysis}, pp. 1--14, 2020, \href{https://doi.org/10.1109/SC41405.2020.00045}{doi:10.1109/SC41405.2020.00045}.

\end{thebibliography}
\end{document}